\documentclass[runningheads]{llncs}
\usepackage[T1]{fontenc}
\usepackage{graphicx}
\usepackage{subfigure}
\usepackage{algpseudocode}
\usepackage{algorithm}
\usepackage{comment}
\usepackage[acronym]{glossaries}
\usepackage{siunitx}
\usepackage{color,soul}
\usepackage{adjustbox}
\usepackage{url}
\usepackage{booktabs}
\usepackage{array}
\usepackage{float}
\usepackage{placeins}
\usepackage{tabularx}

\begin{document}

\title{Machine and Deep Learning for Indoor UWB Jammer Localization}
\titlerunning{Non-Contrastive SSL for NIDS}
%
%
\author{Hamed Fard\inst{1}\orcidID{0009-0007-2365-4313} \and
Mahsa Kholghi\inst{3}\orcidID{0000-0001-6881-1852} \and
Benedikt Groß\inst{1}\orcidID{0000-0002-1112-4426} \and
Gerhard Wunder\inst{1}\orcidID{0009-0001-0850-8816}}
\authorrunning{H. Fard et al.}
%
\institute{Freie Universität Berlin, Germany \\
\email{\{h.habibi.fard,benedikt.gross,g.wunder\}@fu-berlin.de}\\
 Ruhr Universität Bochum Germany\\
\email{\{mahsa.kholghi\}@rub.de}} 
\maketitle              
\begin{abstract}
Ultra-wideband (UWB) localization delivers centimeter-scale accuracy but is vulnerable to jamming attacks, creating security risks for asset tracking and intrusion detection in smart buildings. Although ML and DL methods have improved tag localization, localizing malicious jammers within a single room and across changing indoor layouts remains largely unexplored. Two novel UWB datasets, collected under original and modified room configurations, are introduced to establish comprehensive ML/DL baselines. Performance is rigorously evaluated using a variety of classification and regression metrics. On the source dataset with the collected UWB features, Random Forest achieves the highest F1-macro score of 0.95 and XGBoost achieves the lowest mean Euclidean error of 20.16 cm. However, deploying these source-trained models in the modified room layout led to severe performance degradation, with XGBoost’s mean Euclidean error increasing tenfold to 207.99\,cm, demonstrating significant domain shift. To mitigate this degradation, a domain-adversarial ConvNeXt autoencoder (A-CNT) is proposed that leverages a gradient-reversal layer to align CIR-derived features across domains. The A-CNT framework restores localization performance by reducing the mean Euclidean error to 34.67 cm. This represents a 77\% improvement over non-adversarial transfer learning and an 83\% improvement over the best baseline, restoring the fraction of samples within 30\,cm to 0.56. Overall, the results demonstrate that adversarial feature alignment enables robust and transferable indoor jammer localization despite environmental changes. Code and dataset available at \url{https://github.com/afbf4c8996f/Jammer-Loc}


\end{abstract}

\section{Introduction}
\label{Introduction}

Ultra-wideband (UWB) ranging provides centimeter-scale accuracy for real-time positioning, yet its reliance on short-duration pulses and correlation-based receivers makes it vulnerable to jamming attacks \cite{poturalski2010cicada,yang2024uwbad}. Robust and secure UWB localization is thus critical for applications such as asset tracking in restricted facilities and intrusion detection in smart buildings. Localization methods leveraging angle-of-arrival (AoA), time-of-flight (ToF), and received signal strength indicator (RSSI) have been extensively studied \cite{fischer2025systematic,aditya2018survey,yang2013rssi}.

More recently, machine learning (ML) and deep learning (DL) approaches have been applied to UWB localization tasks, primarily focusing on classifying line-of-sight (LOS) vs non-line-of-sight (NLOS) conditions or regressing coordinates using features derived from channel impulse responses (CIR) \cite{jiang2020uwb,bregar2018improving}. However, these studies exhibit two critical limitations. First, existing research typically targets localization of legitimate tags rather than malicious jamming sources. Second, the challenge of maintaining model performance across varied physical environments has received relatively limited attention within these ML and DL frameworks. Consequently, robust and domain-invariant jammer localization has not been sufficiently explored.

This paper addresses these gaps through systematic investigation. First, performance baselines are established using ML and DL models trained on a newly collected source dataset, demonstrating strong results for both classification and regression. This initial benchmarking is critical, as models that do not achieve satisfactory performance in the source environment might lack the informative feature representations required for successful adaptation to different settings. Given the greater practical importance of accurate jammer localization, regression models are subsequently evaluated on a second dataset to formally quantify performance degradation caused by domain shift. To mitigate this degradation, a domain-adversarial neural network (DANN) framework \cite{ganin2016domain} is adapted, utilizing a denoising ConvNeXt autoencoder \cite{liu2022convnet} to learn compact CIR-derived feature representations and a gradient-reversal layer \cite{ganin2016domain} to promote domain-invariant learning.

The main contributions of this work are:
\begin{enumerate}
\item Introduce two UWB datasets that include diagnostic features and raw CIR taps recorded in the same room on different days, initially and after changing the room setup.
\item Establish comprehensive classification and regression baselines using diagnostic features, supported by automated hyperparameter optimization and interpretability analyses (SHAP \cite{lundberg2017unified}, mutual information, eta-squared ($\eta ^2 $)).
\item Adapt a domain-adversarial ConvNeXt autoencoder framework to align CIR-derived features across source and target domains, mitigating domain shift effects.
\item Compared the proposed domain-adversarial ConvNeXt autoencoder against its non-adversarial counterpart and two unsupervised domain adaptation baselines (CORAL \cite{sun2016return} and MMD \cite{long2015learning}) to quantify the benefits of adversarial alignment. 
\end{enumerate}

The remainder of this paper is organized as follows. Section \ref{sec:related} reviews related work. Section \ref{sec:setup} describes the threat model and experimental setup. Section \ref{sec:methods} 
 presents the baseline methods, feature extraction, hyperparameter optimization, and the domain-adversarial framework. Section \ref{sec:results} reports experimental results. Section \ref{sec:conclusion} concludes and outlines future work.

\section{Related Work}
\label{sec:related}

Prior research on UWB jammer localization has concentrated on detecting and correcting jammed range measurements. For example, \cite{jun2022precise} addresses simulated jamming by classifying corrupted ranges with K-nearest neighbors and correcting them using neural networks. Similarly, \cite{peterseil2023trustworthiness} proposed a trustworthiness score derived from an autoencoder and evaluated robustness under jamming conditions using a publicly available UWB dataset released by those authors. These approaches, however, focus on mitigating specific interference scenarios without addressing generalization to unseen environments.

Domain adaptation techniques have also been explored to tackle distribution shifts between diverse environments or conditions in UWB localization. These methods address either regression tasks (estimating location parameters) or classification tasks (NLOS identification or location prediction). For regression, an adversarial bi-regressor network is utilized, leveraging multi-modal RF data, including UWB, to address layout changes through adversarial training \cite{xia2022adversarial}. In another domain generalization method, regression tasks such as ranging error and AoA estimation are performed by extracting domain-invariant features through domain adversarial training combined with mutual information estimation \cite{xue2025generalization}. For classification tasks, a dual-domain MLP-Mixer framework is applied, employing domain alignment strategies achieved by minimizing a combined loss consisting of classification, weight regularization, and Maximum Mean Discrepancy (MMD) terms \cite{li2025d}. Additionally, CIR-based domain adaptive methods have been proposed for UWB NLOS classification (identification), wherein domain mappings are used to align source and target CIRs by jointly minimizing domain shifts, marginal distribution divergence, and conditional distribution divergence using the MMD criterion \cite{nkrow2025adalos}. However, existing domain adaptation techniques in UWB localization primarily address the task of locating benign tags under distribution shifts or layout changes. To the best of our knowledge, robust localization of UWB jammers under such challenging conditions has not yet been specifically explored in existing literature.


\section{Threat Model and Experimental Setup}
\label{sec:setup}
\subsection{Threat Model}

The threat model assumes an attacker seeking to disrupt UWB-based localization. While UWB communication is often considered jamming-resistant due to its wide bandwidth, this assumption can be invalidated by practical hardware limitations. The DWM3001CDK development kits \footnote{Documentation available at \url{https://www.qorvo.com/products/p/DWM3001CDK#documents}.} used in this work, for example, support only UWB channels 5 and 9. An adversary aware that all anchor nodes operate on the 6.5~GHz front end (channel 5) can launch a highly effective, targeted jamming attack.

The adversary is assumed to employ a DW3000-series transceiver, identical to those of the receivers (anchors), configured to transmit continuous UWB frames at maximum regulatory power. This is implemented via a simple firmware loop in C using the Decawave API. 
The attacker gains knowledge of channel parameters, including data rate, preamble length, and synchronization codes, through a packet-sniffing algorithm as presented in \cite{yang2024uwbad}. The attacker possesses no specialized hardware advantages or advanced capabilities such as beamforming. Unlike conventional UWB-based tag localization, which assumes interference-free reception and relies on clean signals for position estimation, jammer localization is more challenging due to the reduced effective signal-to-noise ratio and the need to distinguish overlapping transmissions. Under these conditions, the goal of the localization system is to determine the jammer’s two-dimensional position, given that each anchor node receives a superposition of legitimate tag pulses and continuous jamming frames.

\subsection{Experimental Setup}

\begin{figure}
    \centering
    \includegraphics[width=0.5\linewidth]{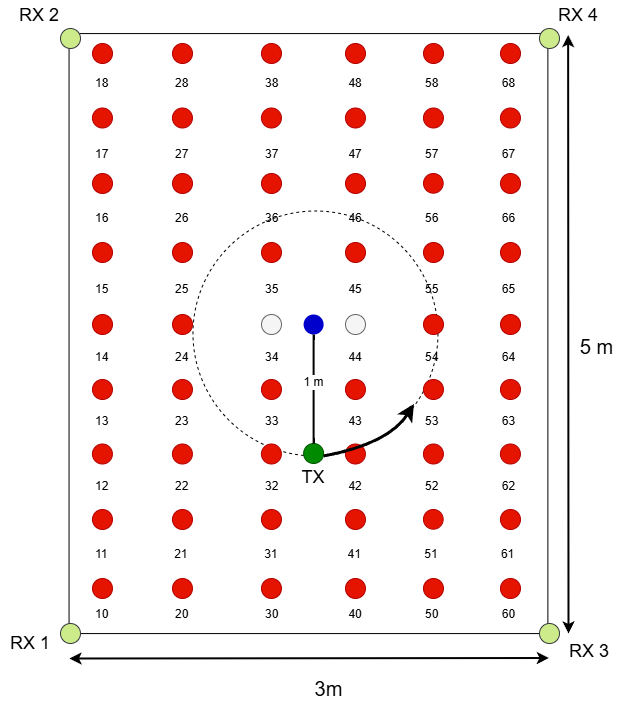}\hfill
    \raisebox{1.5em}{\includegraphics[width=0.5\linewidth]{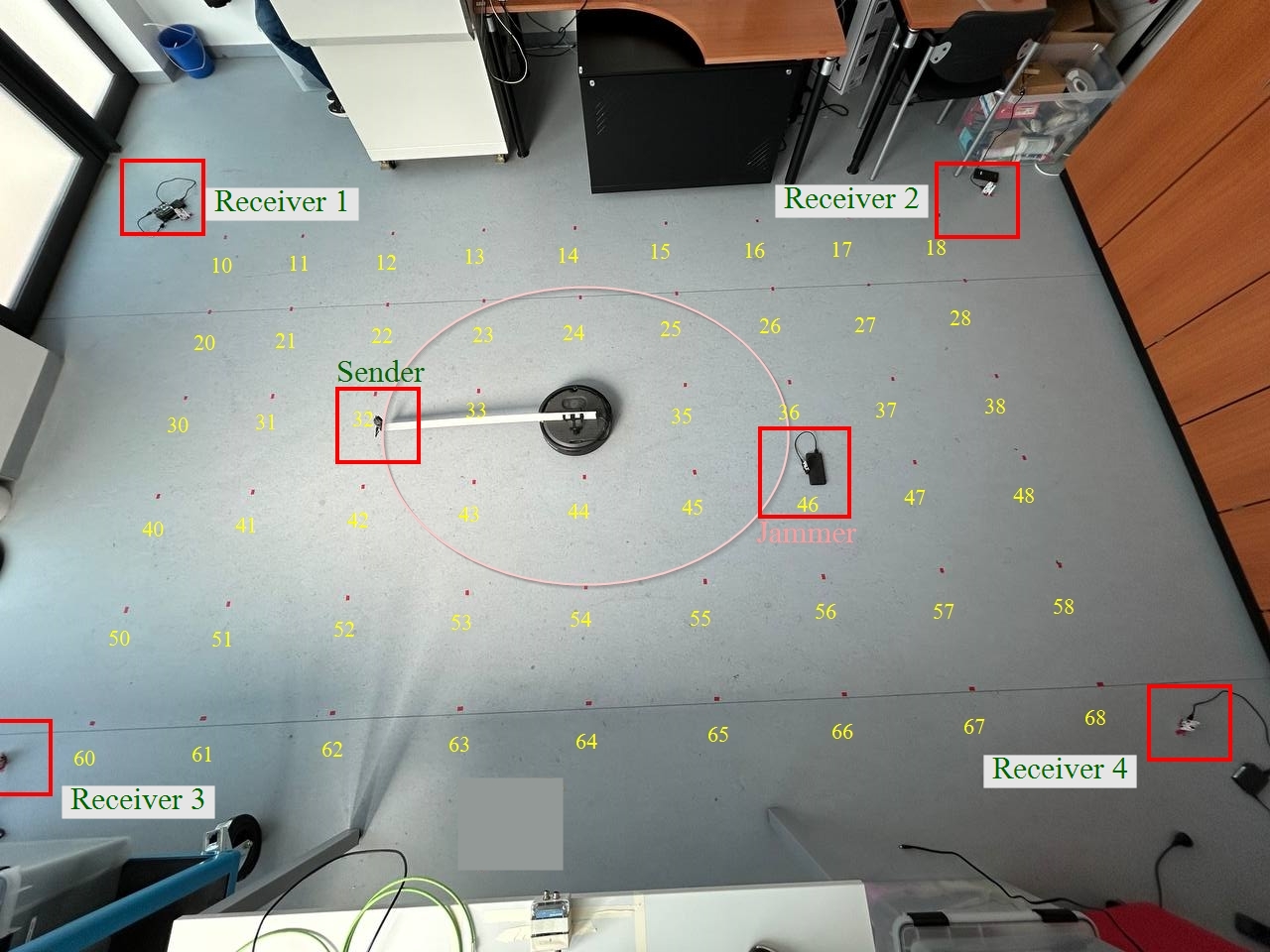}}
    \caption{Left: Schematic layout of the experimental environment showing the positions of the four UWB receivers (RX 1–4), the transmitter (TX) mounted on a TurtleBot robot with a 1\,m extension arm, and the 52 distinct jammer positions (red dots) within a 3\,m×5\,m area. Right: The actual indoor setup, with the sender (TX), receivers (RX), and marked jammer locations on the floor corresponding to the schematic.}
    \label{fig:positions}
\end{figure}

\noindent\textbf{Environment and Configurations.}
Experiments were conducted in an indoor laboratory over a 3\,m~$\times$~5\,m test grid. To study the effect of domain shift, data was collected in two distinct environmental configurations. The source domain consisted of the initial laboratory environment with standard furniture, where data was collected at 52 predefined jammer locations. For the target domain, the environment was perturbed by moving two desks, one chair, and several smaller objects outside the test grid. Data was then collected at 16 new, randomly selected jammer locations to simulate deployment in a different layout. The jammer positions for the source domain are illustrated in Figure~\ref{fig:positions}, while those for the target domain appear on the right side of Figure~\ref{fig:k-means}.

\noindent\textbf{Hardware Components.}
Four UWB receivers were positioned at the corners of the test grid. A legitimate UWB transmitter (tag) was mounted on a TurtleBot mobile robot via a 1\,m extension arm. A separate DW3000-series transceiver, identical to the receivers, served as the interference source (jammer). Both the legitimate tag and the jammer operated in High Pulse Repetition (HPR) mode at standardized power levels.

\noindent\textbf{Data Collection Protocol.}
The data collection followed a precise, automated protocol for each jammer position. First, the jammer was placed at a predefined $(x,\,y)$ location on the grid and began transmitting continuously. Next, to simulate a dynamic operational environment with varying signal paths, the TurtleBot carrying the legitimate tag executed a full 360° rotation at the center of the grid while emitting packets at 10\,ms intervals. During this rotation, each of the four receiver nodes recorded per-corner diagnostic readings from the DW3000 registers and the raw CIR taps. The collected measurements were subsequently annotated with the ground-truth $(x,\,y)$ coordinates of the stationary jammer. This entire process was repeated for all 52 jammer positions in the source domain and all 16 positions in the target domain, resulting in two distinct datasets for training and evaluation. Data collection resulted in $461,795$ samples for the source domain and $28,793$ samples for the target domain. The dataset will be made publicly available.


\section{Methodology}
\label{sec:methods}

\subsection{Feature Engineering and Data Preprocessing}
\label{subsec:feature_engineering}

For baseline experiments on the source dataset, numeric diagnostic features from the DW3000 transceivers (described in Table~\ref{tab:features}) were normalized using a standard scaler fitted to the training set. Scaler parameters from training were directly applied to validation, test, and target datasets. For the DANN pipeline, complex CIR samples were separately preprocessed. Each sample, originally comprising 300 taps, was converted into magnitude, sine phase, and cosine phase, truncated to the first 100 taps, and normalized using a standard scaler fitted jointly to source and target datasets. Truncation to 100 taps was applied after preliminary experiments indicated that tap sequences longer than 100 taps degraded model performance and introduced excessive noise. No other feature preprocessing techniques were applied beyond those described above, thereby preserving the generality of the pipeline.

\setlength{\tabcolsep}{6pt}  

\begin{table}[H]
  \centering
  \caption{DW3000 Feature Descriptions}
  \label{tab:features}
  \begin{adjustbox}{%
      width=0.65\textwidth,         
      max height=0.3\textheight,   
      keepaspectratio              
    }
    \begin{tabular}{@{} l p{0.50\linewidth} @{}}
      \toprule
      \textbf{Feature}
        & \multicolumn{1}{>{\centering\arraybackslash}p{0.58\linewidth}}{\textbf{Description}} \\
      \midrule
      PHE         & Diagnostic reading for PHY‐header error detection.            \\
      RSL         & Diagnostic reading for Reed–Solomon decoding failures.         \\
      CRCG        & Diagnostic reading for CRC/FCS integrity passes.               \\
      CRCB        & Diagnostic reading for CRC/FCS integrity failures.             \\
      PREJ        & Diagnostic reading for preamble rejection occurrences.         \\
      RSSI        & Received signal strength (in dBm) as reported by the DW3000 diagnostics. \\
      IpatovPeak  & Amplitude of the strongest path detected in the channel impulse response. \\
      IpatovPower & Total energy measured over the channel impulse response window.  \\
      IpatovF1    & Early‐time CIR energy measurement used to help locate the first‐path. \\
      IpatovF2    & Mid‐time CIR energy measurement used to refine the first‐path timing estimate.      \\
      IpatovF3    & Late‐time CIR energy measurement used to further sharpen the first‐path timing estimate. \\
      CMPLX\_CIR  & Raw complex CIR vector of 300 taps.                             \\
      \bottomrule
    \end{tabular}
  \end{adjustbox}
\end{table}

\subsection{Baseline Models: Machine and Deep Learning Approaches}
\label{subsec:baseline_models}

Baseline models included Random Forest (RF), XGBoost (XGB), and K-Nearest Neighbors (KNN), as well as three deep learning models: SimpleNN (fully-connected residual network \cite{he2016deep}), ConvMixer1D (CNN using depthwise convolutions \cite{chollet2017xception} to mix channel-wise features), and Transformer Tabular (encoder-only transformer \cite{vaswani2017attention} without positional encoding). To ensure fair comparisons, hyperparameters for each ML and DL model were separately optimized for classification and regression tasks using Bayesian optimization with Optuna \cite{akiba2019optuna}. Optimal hyperparameters, such as learning rates, network size, number of epochs, and other training details, are determined by the code in our public repository.

\subsection{Adversarial Domain Adaptation}
\label{subsec:domain_adaptation}
\noindent\textbf{Motivation.}
Domain shift refers to differences between source and target domain data distributions, which typically cause significant performance degradation when models trained in one domain are applied to another \cite{farahani2021brief,singhal2023domain}. Such discrepancies challenge the robustness and generalization of deployed models. Unsupervised Domain Adaptation (UDA) methods tackle this issue by transferring knowledge from labeled source domain data to unlabeled or sparsely labeled target domain data \cite{liu2022deep,wilson2020survey}. Domain-adversarial approaches specifically aim at learning domain-invariant representations, improving model adaptability and generalization across domains \cite{hassanpour2023survey}. A significant performance degradation was observed in this study when models trained and validated on the source dataset were evaluated on the target dataset, as will be shown in Section \ref{sec:results}. To address this challenge, a domain-adversarial neural network (DANN) utilizing a denoising ConvNeXt autoencoder is employed. The autoencoder explicitly learns robust, domain-invariant representations, reducing domain discrepancy between source and target domains.

\noindent\textbf{Model Architecture.}
The autoencoder is a compact, lightweight ConvNeXt-based architecture designed explicitly for processing CIR data. It takes as input 100 CIR taps with three channels (magnitude, sine phase, cosine phase). The encoder progressively compresses the input via three convolutional downsampling blocks, increasing channels from 3 to 128, with intermediate ConvNeXt residual blocks featuring layer normalization, depthwise convolutions, and GELU activations. Gaussian noise ($\sigma = 0.6$) is explicitly injected into encoder activations to promote robustness to input perturbations and prevent trivial identity mappings. The decoder symmetrically mirrors the encoder structure, utilizing transpose convolutional layers to restore the input dimensionality back to its original form. In total, the autoencoder comprises 782,211 parameters, making it efficient and suitable for resource-constrained scenarios.

\noindent\textbf{Gradient Reversal Layer (GRL).} 
The GRL \cite{ganin2016domain} is inserted between the encoder and the domain classifier in the denoising ConvNeXt autoencoder pipeline. During the forward pass it passes the encoder’s feature embeddings unchanged, but in backpropagation it multiplies the domain classifier’s gradients by $-\lambda$. This gradient inversion encourages the encoder to learn features that are uninformative for domain discrimination, thereby aligning source and target feature distributions.

\noindent\textbf{Domain Classifier.} The domain classifier distinguishes source from target domain features, enabling adversarial alignment. Its architecture comprises two fully connected layers (128 and 64 units, respectively) with ReLU activations, followed by an output layer with sigmoid activation to predict domain labels. 

\noindent\textbf{Regression Head.} The regression head predicts continuous target variables (i.e., coordinates). It is implemented as a single fully-connected linear layer that maps the shared encoder representation directly to the regression targets without intermediate activations.

\noindent\textbf{Training Procedure.} 
Training consists of three phases:   

\noindent\textbf{Autoencoder Pre-training.} The ConvNeXt autoencoder is pre-trained exclusively on unlabeled source-domain data. This unsupervised training minimizes reconstruction loss between Gaussian-noise-injected inputs ($\sigma= 0.6 $) and actual clean signals. Training occurs over $30$ epochs with an initial learning rate of $1e-3$, progressively adjusted using linear warmup followed by cosine annealing.

\noindent\textbf{Joint Adversarial Alignment.} The pre-trained autoencoder undergoes adversarial alignment using data without class labels from both source and target domains, with domain labels used only for adversarial training. The model jointly minimizes reconstruction loss on target data and domain-classification loss via a gradient reversal layer, with the reversal strength parameter ($\lambda$) following a sigmoid schedule from $0.05$ to $0.2$ over $40$ epochs. Domain-classifier performance is monitored using AUC, which initially increases to $0.66$ and then gradually decreases as alignment progresses. Early stopping is triggered at epoch $18$ based on AUC stagnation, with the score reaching $0.5145$ and indicating convergence toward domain confusion, while reconstruction loss remains stable.

\begin{figure}[htbp]
  \centering
  \subfigure[]{%
    \includegraphics[width=0.48\textwidth]{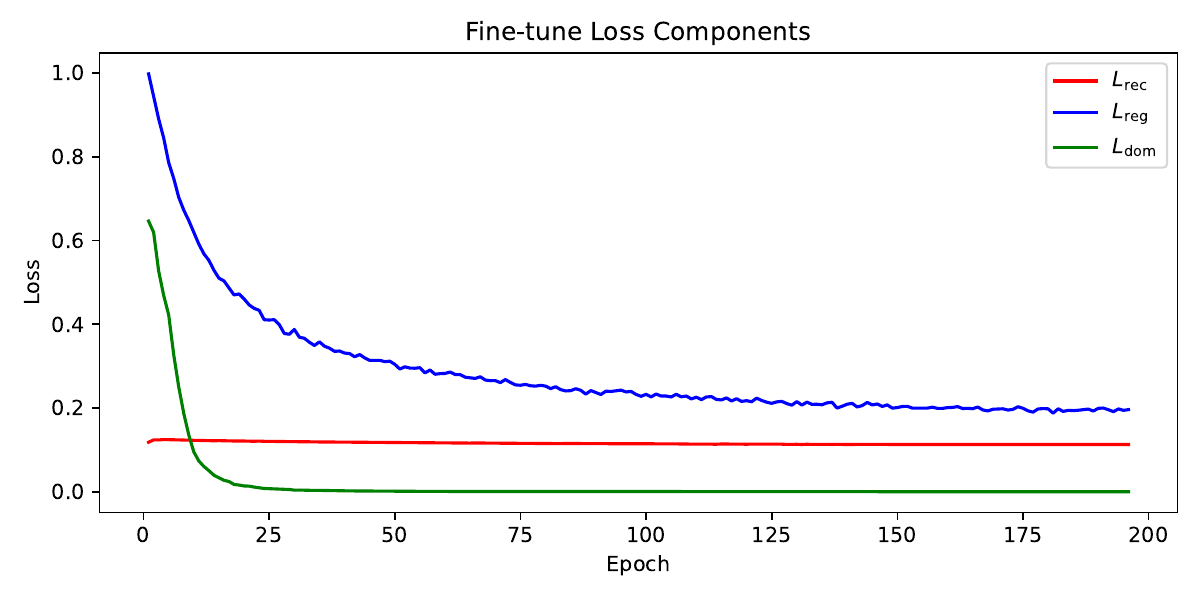}%
    \label{fig:ft-loss}%
  }\hfill
  \subfigure[]{%
    \includegraphics[width=0.48\textwidth]{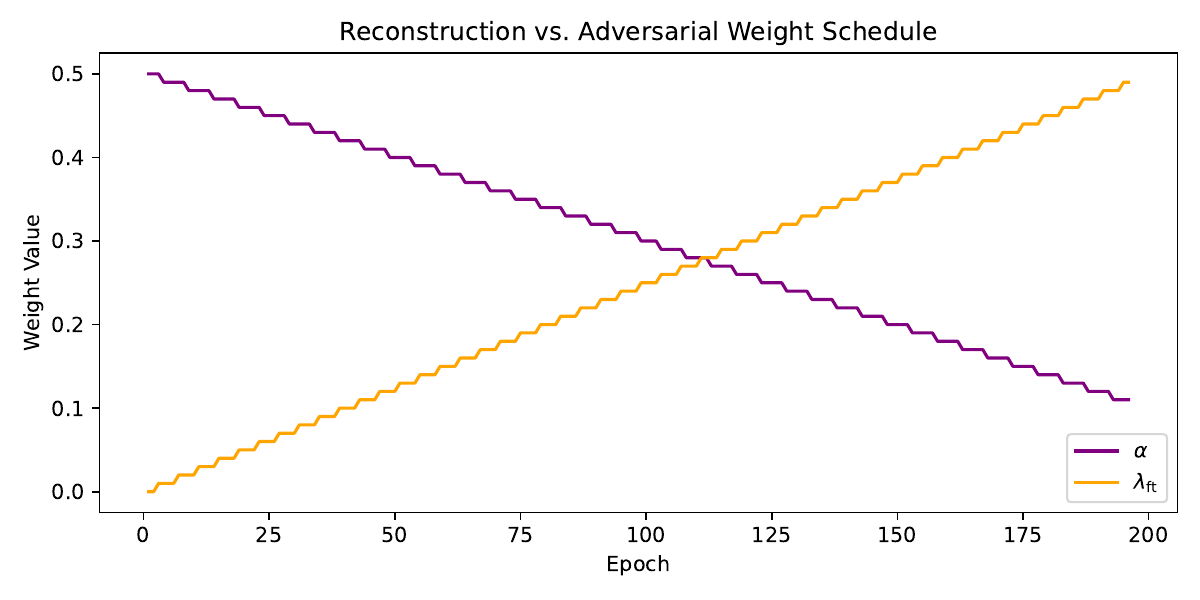}%
    \label{fig:ft-weights}%
  }\hfill

  \caption{Fine-tuning convergence over 200 epochs: (a) loss components \(L_{\mathrm{rec}}\), \(L_{\mathrm{reg}}\), \(L_{\mathrm{dom}}\); (b) weight schedules for \(\alpha\) and \(\lambda_{\mathrm{ft}}\)}
  \label{fig:fine_tune}
\end{figure}

\noindent\textbf{Fine-Tuning.} Fine-tuning employs labeled target-domain data for supervised regression of spatial coordinates. Only the autoencoder’s final encoder stage, decoder layers, final convolution layer, and the regression head are unfrozen. The domain classifier remains active during fine-tuning, acting as a regularizer to maintain domain-invariant feature representations. Training minimizes a combined loss composed of reconstruction loss, regression loss and and adversarial domain loss. Over 200 epochs, hyperparameters $\alpha$ and $\lambda_{ft}$ are progressively adjusted (from $0.5$ to $0.1$ and from $0.0$ to $0.5$, respectively), while $\beta$ remains constant at $1.0$. Cosine annealing schedulers are used for both the regression head and autoencoder learning rates. Training continues until performance plateaus, explicitly evaluated on a dedicated hold-out set for unbiased assessment of adaptation effectiveness. The key quantities described above for the fine-tuning phase are plotted in Figure~\ref{fig:fine_tune}. In Figure~\ref{fig:fine_tune}(a), the reconstruction loss \(L_{\mathrm{rec}}\), regression loss \(L_{\mathrm{reg}}\) and adversarial domain loss \(L_{\mathrm{dom}}\) are shown over 200 epochs to illustrate their evolution during fine-tuning. Figure~\ref{fig:fine_tune}(b) displays the annealing schedules for the reconstruction weight \(\alpha\) and adversarial weight \(\lambda_{ft}\).

\noindent\textbf{Formalization of Loss Components.}  
To summarize the losses utilized at each training phase mathematically, the following equations define reconstruction, regression, and adversarial losses employed (Eqs.\ref{eq:l_rec}-\ref{eq:l_ft}). In these equations, $x_i^t$ denotes unlabeled samples from the target domain, $x_j^s$ denotes unlabeled samples from the source domain, and $x_i$ denotes labeled target-domain samples used for supervised fine-tuning. The terms $N_t$, $N_s$, and $N_c$ represent the respective number of samples in these sets. $AE_{\theta}(\cdot)$ is the ConvNeXt autoencoder, and $f_{\theta}(\cdot)$ indicates the autoencoder's learned feature extractor (bottleneck representation). $D_{\phi}(\cdot)$ represents the domain classifier, while $R_{\psi}(\cdot)$ is the regression head predicting spatial coordinates. Finally, $\mathrm{GRL}_{\lambda}(\cdot)$ denotes the Gradient Reversal Layer, which acts as an identity function during forward propagation and reverses gradients scaled by $-\lambda$ during backpropagation.

\begin{align}
\mathcal{L}_{\mathrm{rec}} &= \frac{1}{N_t}\sum_{i=1}^{N_t}\bigl\lVert AE_{\theta}(x_i^t) - x_i^t \bigr\rVert_{2}^{2}, \label{eq:l_rec}\\[6pt]
\mathcal{L}_{\mathrm{dom}} &= \frac{1}{N_s + N_t}\Biggl[ 
    \sum_{j=1}^{N_s}\mathrm{BCE}\bigl(D_{\phi}(\mathrm{GRL}_{\lambda}(f_{\theta}(x_j^s))), 0\bigr) \nonumber \\[4pt]
    &\hspace{1.7cm}+ \sum_{i=1}^{N_t}\mathrm{BCE}\bigl(D_{\phi}(\mathrm{GRL}_{\lambda}(f_{\theta}(x_i^t))), 1\bigr)\Biggr], \label{eq:l_dom}\\[6pt]
\mathcal{L}_{\mathrm{adapt}} &= \mathcal{L}_{\mathrm{rec}} + \lambda\,\mathcal{L}_{\mathrm{dom}}, \label{eq:l_adapt}\\[10pt]
\mathcal{L}_{\mathrm{rec}}^{\mathrm{ft}} &= \frac{1}{N_c}\sum_{i=1}^{N_c}\bigl\lVert AE_{\theta}(x_i) - x_i \bigr\rVert_{2}^{2}, \label{eq:l_rec_ft}\\[6pt]
\mathcal{L}_{\mathrm{reg}} &= \frac{1}{N_c}\sum_{i=1}^{N_c}\bigl\lVert R_{\psi}(f_{\theta}(x_i)) - y_i \bigr\rVert_{2}^{2}, \label{eq:l_reg}\\[6pt]
\mathcal{L}_{\mathrm{dom}}^{\mathrm{ft}} &= \frac{1}{N_c}\sum_{i=1}^{N_c}\mathrm{BCE}\bigl(D_{\phi}(\mathrm{GRL}_{\lambda_{\mathrm{ft}}}(f_{\theta}(x_i))), 0\bigr), \label{eq:l_dom_ft}\\[6pt]
\mathcal{L}_{\mathrm{ft}} &= \alpha\,\mathcal{L}_{\mathrm{rec}}^{\mathrm{ft}} + \beta\,\mathcal{L}_{\mathrm{reg}} + \lambda_{\mathrm{ft}}\,\mathcal{L}_{\mathrm{dom}}^{\mathrm{ft}}. \label{eq:l_ft}
\end{align}

\textbf{Evaluation metrics.} Experiments report total runtime in minutes (Time) and the following regression performance metrics: mean, median, and 90th percentile (P90) of Euclidean distance errors between predicted and actual spatial coordinates (in centimeters), the fraction of predictions within 30 cm (F$_{\le30\mathrm{cm}}$), and root mean squared error (rmse), mean absolute error (mae), and coefficient of determination ($R^2$) computed separately for each spatial axis $(x,\,y)$ to quantify anisotropic prediction quality.

\subsection{Implementation and Computational Resources}
\label{subsec:implementation}
Experiments ran on a 64-bit Debian GNU/Linux 12 workstation with an Intel Core i9-10980XE CPU, 32 GB RAM, and an NVIDIA RTX A5500 GPU (24 GB VRAM), using PyTorch.

\section{Experiments and Results}
\label{sec:results}

\begin{figure}[!htbp]
  \centering
  \subfigure[SimpleNN]{%
    \includegraphics[width=0.25\textwidth]{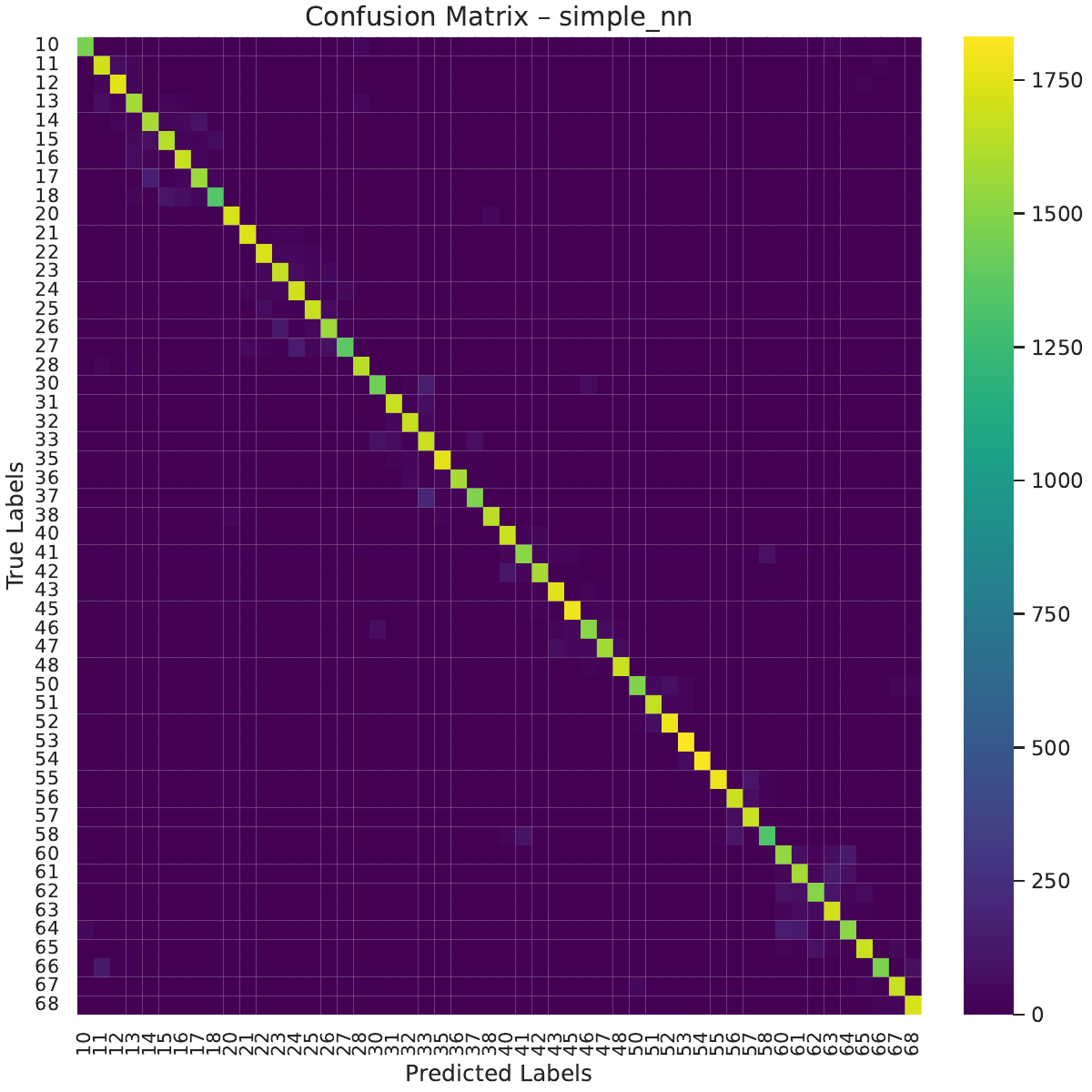}%
    \label{fig:cm1}%
  }\hfill
  \subfigure[ConvMixer1D]{%
    \includegraphics[width=0.25\textwidth]{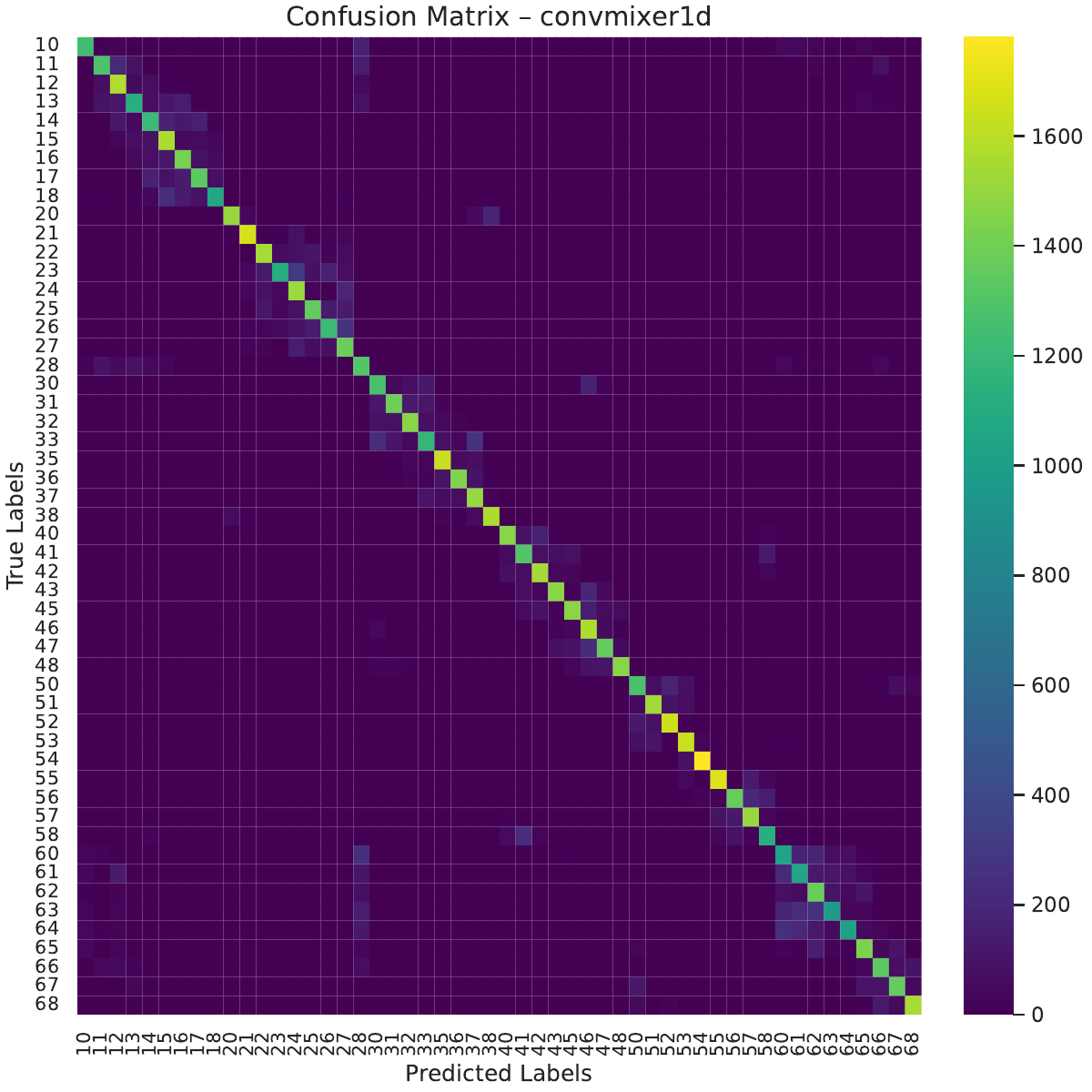}%
    \label{fig:cm2}%
  }\hfill
  \subfigure[Transformer]{%
    \includegraphics[width=0.25\textwidth]{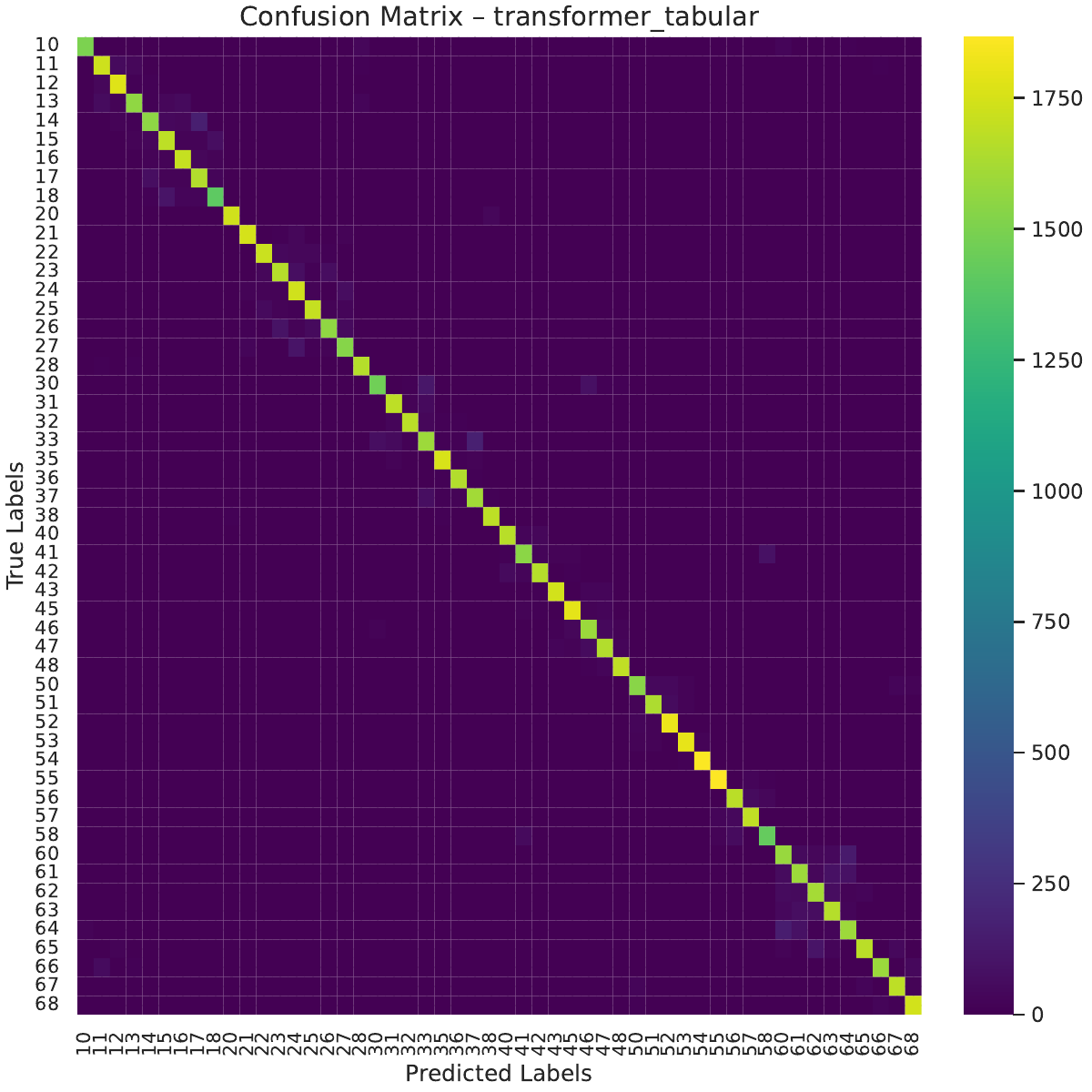}%
    \label{fig:cm3}%
  }
  \caption{Confusion matrices of the three deep learning models evaluated on the source dataset (52 classes).}
  \label{fig:cms}
\end{figure}

\noindent\textbf{Classification on source dataset.}
The goal of this experiment is to evaluate the classification performance of various ML and DL models when trained, validated, and tested solely on the source dataset. Table~\ref{tab:classification_results} shows that RF achieved an accuracy of $0.95$ and an F1-macro of $0.95$ in under 3 minutes, followed by XGB with 0.94 accuracy and F1-macro in the same time frame. The Transformer model reached $0.93$ accuracy in $34$ minutes, S\_NN obtained $0.91$ in $11$ minutes, and CM1D reached $0.78$ in $13$ minutes. By contrast, KNN achieved only 0.50 accuracy, likely because the large number of classes ($52$) dilutes neighbor density in the feature space. The close agreement between F1-macro and F1-weighted scores indicates balanced performance across all classes. These results demonstrate that classical ML models achieve competitive classification performance with substantially lower computational cost. Furthermore, the high and balanced scores confirm the utility of the diagnostic features for this task. Figure~\ref{fig:cms} shows the confusion matrices for the three DL models.

\begin{table}[H]
  \centering
  \caption{Classification performance of models on source dataset (52 classes).}
  \label{tab:classification_results}
  \resizebox{0.7\textwidth}{!}{%
    \begin{tabular}{@{} l
                     S[table-format=1.2]
                     S[table-format=1.2]
                     S[table-format=1.2]
                     S[table-format=1.2]
                     S[table-format=1.2]
                     S[table-format=2.2]
                     @{}}
      \toprule
      \textbf{Model}
        & {\textbf{Accuracy}}
        & {\textbf{F1-macro}}
        & {\textbf{F1-weighted}}
        & {\textbf{Precision}}
        & {\textbf{Recall}}
        & {\textbf{Time}} \\
      \midrule
      CM1D
        & 0.7797 & 0.7800 & 0.7797 & 0.7877 & 0.7803 & 13 \\
      S\_NN
        & 0.9144 & 0.9146 & 0.9143 & 0.9163 & 0.9142 & 11 \\
      Trans
        & 0.9310 & 0.9313 & 0.9310 & 0.9318 & 0.9312 & 34 \\
      XGB
        & 0.9381 & 0.9383 & 0.9380 & 0.9385 & 0.9383 & \textbf{3.00} \\
      RF
        & \textbf{0.95} & \textbf{0.95} & \textbf{0.95}
        & \textbf{0.95} & \textbf{0.95} & \textbf{3.00} \\
      KNN
        & 0.5041 & 0.5046 & 0.5044 & 0.5056 & 0.5040 & 6.00 \\
      \bottomrule
    \end{tabular}%
  }
\end{table}

\textbf{Regression on source dataset.}
The goal of this experiment is to assess the regression performance of various ML and DL models for spatial localization when trained, validated, and tested solely on the source dataset. Table~\ref{tab:regression_results} shows that XGB achieved the lowest mean error (20.16\,cm) and median error (10.54\,cm) with \(F_{\le30\mathrm{cm}}=0.80\) in 2\,minutes. CM1D and S\_NN both incurred mean errors near 28\,cm and \(F_{\le30\mathrm{cm}}<0.75\), indicating only moderate localization accuracy from these DL architectures. RF exhibited unbalanced axis errors with \(\mathrm{rmse}_X=3.23\)\,cm and \(\mathrm{rmse}_Y=59.90\)\,cm leading to a mean error of 43.15\,cm and \(F_{\le30\mathrm{cm}}=0.48\). The 90th-percentile error (P90) further underscores XGB’s robustness (50.84\,cm) compared to RF (100.60\,cm) and the DL baselines. Across all models, \(R^2_X\ge0.99\) confirms a strong fit along the X axis whereas lower \(R^2_Y\) values reflect greater variance along the Y axis. These results suggest that in the source domain classical ML models, particularly XGB, outperform the deep architectures in both accuracy and robustness to outliers and do so with comparable or lower training times.

\begin{table}[H]
  \centering
   \footnotesize
  \renewcommand{\arraystretch}{1.1}   
  \setlength{\tabcolsep}{1.5pt}         
  \caption{Regression performance metrics for localization on the source dataset.}
  \label{tab:regression_results}
  \begin{tabular}{@{} l
                  S[table-format=2.2]
                  S[table-format=2.2]
                  S[table-format=2.2]
                  S[table-format=2.2]
                  S[table-format=1.2]
                  S[table-format=1.2]
                  S[table-format=2.2]
                  S[table-format=2.2]
                  S[table-format=3.2]
                  S[table-format=1.2]
                  S[table-format=2.0]
                  @{}}
    \toprule
      \textbf{Model}
        & {\textbf{rmse$_X$}}
        & {\textbf{rmse$_Y$}}
        & {\textbf{mae$_X$}}
        & {\textbf{mae$_Y$}}
        & {\textbf{$R^2_X$}}
        & {\textbf{$R^2_Y$}}
        & {\textbf{mean}}
        & {\textbf{med}}
        & {\textbf{P90}}
        & {\textbf{F$_{\le30\mathrm{cm}}$}}
        & {\textbf{Time}} \\
    \midrule
    CM1D
      &  7.40 & 45.31 &  4.04 & 27.81
      & $\mathbf{0.99}$ & 0.87
      & 28.79 & 15.74 & 73.04
      & 0.71 & 10 \\
    S\_NN
      & 16.80 & 45.40 &  5.97 & 26.08
      & 0.96 & 0.87
      & 28.23 & 13.37 & 73.35
      & 0.74 &  4 \\
    Trans
      & 15.83 & 46.60 &  7.14 & 29.48
      & 0.96 & 0.87
      & 31.86 & 17.92 & 76.46
      & 0.68 & 17 \\
    XGB
      &  4.51 & $\mathbf{33.50}$ &  2.43 & $\mathbf{19.52}$
      & $\mathbf{0.99}$ & $\mathbf{0.93}$
      & $\mathbf{20.16}$ & $\mathbf{10.54}$ & $\mathbf{50.84}$
      & $\mathbf{0.80}$ &  $\mathbf{2}$ \\
    RF
      &  $\mathbf{3.23}$ & 59.90 &  $\mathbf{0.73}$ & 43.10
      & $\mathbf{0.99}$ & 0.78
      & 43.15 & 31.34 &100.60
      & 0.48 &  $\mathbf{2}$ \\
    \bottomrule
  \end{tabular}
  \normalsize
\end{table}

\textbf{Feature interpretability.}
The goal of this analysis is to identify which diagnostic readings influence localization error. Table~\ref{tab:feature_importance} gives four importance measures: SHAP, XGBoost gain, mutual information and eta-squared. A mean rank across these measures shows the consensus ordering. The top four features in this ranking are RSL, PHE, PREJ and CRCB. The Ipatov-derived features and RSSI have lower importance. These results confirm the value of the core diagnostic features for localization. They also suggest that the lower ranking features could be removed in future work to reduce model complexity.

\begin{table}[H]
  \caption{Feature‐level importance metrics for the XGBoost regressor predicting Euclidean distance. Lower mean‐rank indicates higher feature importance.}
  \label{tab:feature_importance}
  \centering
  \small
  \setlength{\tabcolsep}{4pt}           
  \renewcommand{\arraystretch}{0.95}     
  \resizebox{0.75\columnwidth}{!}{%
    \begin{tabular}{l
                    c@{\hspace{1em}}
                    c@{\hspace{1em}}
                    c@{\hspace{1em}}
                    c@{\hspace{1em}}
                    c}
      \toprule
      \textbf{Feature}       & \textbf{SHAP} & \textbf{XGB Imp.} & \textbf{Mutual Info} & \textbf{Eta} & \textbf{Mean Rank} \\
      \midrule
      RSL   & 39.72   & 0.415  & 2.410  & 0.366  & 1.00  \\
      PHE   & 26.20   & 0.232  & 2.197  & 0.268  & 2.25  \\
      PREJ  & 27.34   & 0.124  & 1.297  & 0.148  & 3.25  \\
      CRCB  &  8.03   & 0.209  & 1.080  & 0.250  & 3.50  \\
      CRCG  &  3.90   & 0.009  & 0.082  & 0.066  & 5.00  \\
      RSSI              &  0.95   & 0.006  & 0.009  & 0.017  & 7.00  \\
      IpatovPeak        &  0.45   & 0.001  & 0.010  & 0.045  & 7.25  \\
      IpatovF3          &  0.35   & 0.002  & 0.002  & 0.024  & 8.00  \\
      IpatovPower       &  0.29   & 0.001  & 0.002  & 0.018  & 9.00  \\
      IpatovF1          &  0.30   & 0.001  & 0.007  & 0.013  & 9.00  \\
      IpatovF2          &  0.19   & 0.001  & 0.002  & 0.016  & 10.75 \\
      \bottomrule
    \end{tabular}%
  }
\end{table}
\begin{table}[H]
  \centering
  \scriptsize
  \renewcommand{\arraystretch}{1.1}
  \setlength{\tabcolsep}{5pt}
  \caption{Localization performance on the target dataset for models trained exclusively on the source domain and for those employing domain adaptation.}
  \label{tab:regression_results_target}
  \begin{tabular}{@{} l
                  S[table-format=3.2]
                  S[table-format=3.2]
                  S[table-format=3.2]
                  S[table-format=3.2]
                  S[table-format=1.2]
                  S[table-format=1.2]
                  S[table-format=3.2]
                  S[table-format=3.2]
                  S[table-format=3.2]
                  S[table-format=1.2]
                  @{}}
    \toprule
    \textbf{Model}
      & {\textbf{rmse$_X$}} & {\textbf{rmse$_Y$}} & {\textbf{mae$_X$}} & {\textbf{mae$_Y$}}
      & {\textbf{$R^2_X$}}  & {\textbf{$R^2_Y$}}  & {\textbf{mean}}
      & {\textbf{med}}      & {\textbf{P90}}      & {\textbf{F$_{\le30\mathrm{cm}}$}} \\
    \midrule
    CM1D    & 127.18 & 200.91 & 101.56 & 165.20 & -1.24 & -1.37 & 208.95 & 199.14 & 376.82 & 0.02 \\
    S-NN    & 140.04 & 223.63 & 117.62 & 194.18 & -1.71 & -1.94 & 241.75 & 243.29 & 393.52 & 0.02 \\
    Trans   & 142.08 & 208.93 & 116.01 & 174.42 & -1.79 & -1.57 & 225.22 & 221.44 & 381.16 & 0.03 \\
    XGB     & 128.16 & 193.96 & 103.90 & 161.15 & -1.27 & -1.21 & 207.99 & 193.57 & 361.27 & 0.03 \\
    RF      & 125.86 & 169.46 & 102.69 & 139.26 & -1.19 & -0.69 & 188.35 & 175.14 & 344.50 & 0.00 \\
    \cline{1-11}
    Coral   & 102.34 & 165.33 & 84.44  & 137.40 & -0.45 & -0.61 & 173.90 & 166.55 & 300.60 & 0.02 \\
    MMD     & 100.39 & 163.37 & 82.69  & 135.71 & -0.39 & -0.57 & 171.59 & 163.27 & 296.43 & 0.02 \\
    CNT     &  85.51 & 140.38 &  74.25 & 117.41 & -0.01 & -0.16 & 148.02 & 142.61 & 249.05 & 0.03 \\
    A-CNT   &  $\mathbf{24.74}$ &  $\mathbf{38.63}$ &  $\mathbf{16.60}$ &  $\mathbf{27.54}$ &  $\mathbf{0.92}$ &  $\mathbf{0.91}$ & $\mathbf{34.67}$ &  $\mathbf{26.62}$ &  $\mathbf{69.03}$ & $\mathbf{0.56}$ \\
    \cline{1-11}
    \bottomrule
  \end{tabular}
  \normalsize
\end{table}
\FloatBarrier

\textbf{Domain shift on target dataset.}
In this section, the robustness of source-trained models is evaluated under the new room layout. Models trained on the source dataset suffer severe performance degradation when evaluated on the target dataset after the room layout is changed. This domain shift is also visualized in Figure \ref{fig:appendix-domain-shift} in the Appendix, where per-tap differences between source and target features are measured using the Wasserstein distance \cite{peyre2019computational} and the absolute difference in normalized means. Table~\ref{tab:regression_results_target} shows that XGB’s mean error rose from 20.16\,cm to 207.99\,cm and \(F_{\le30\mathrm{cm}}\) fell from 0.80 to 0.03. CM1D and S\_NN exhibited similar failures with mean errors above 200\,cm and negative axis-wise \(R^2\) values indicating performance below that of a constant-mean predictor. These results confirm the severity of the domain shift induced by layout change.

\textbf{Domain adaptation.}
To compare adaptation strategies using CIR features, classical UDA methods (CORAL and MMD) were compared against a non-adversarial method (CNT) and its adversarial variant (A-CNT). While all four methods shared the same encoder, their overall architectures differed: CORAL and MMD used encoder-only models with feature alignment, whereas CNT and A-CNT employed the same autoencoder architecture. The only distinction between CNT and A-CNT was the use of gradient reversal for adversarial alignment in the latter. CORAL and MMD reduced mean error relative to source-only training (e.g., to 173.90\,cm with CORAL), while CNT achieved further improvement with 148.02\,cm. This suggests that the compact representations learned by the autoencoder may generalize better across domains than shallow feature alignment alone. A-CNT achieved the best performance, with a mean error of 34.67\,cm and R² scores of 0.92 and 0.91. These results indicate that classical UDA methods offer benefits over source-only training, but adversarial domain alignment yields significantly greater gains. Furthermore, to qualitatively assess the effect of adversarial domain adaptation, t-SNE plots are provided in the appendix (Figure \ref{fig:Alignment}), visualizing the A-CNT model's feature representations both before and after applying the GRL.

\begin{figure}[H]
  \centering
  \includegraphics[%
    width=0.60\textwidth,%
    height=0.30\textheight,%
    keepaspectratio%
  ]{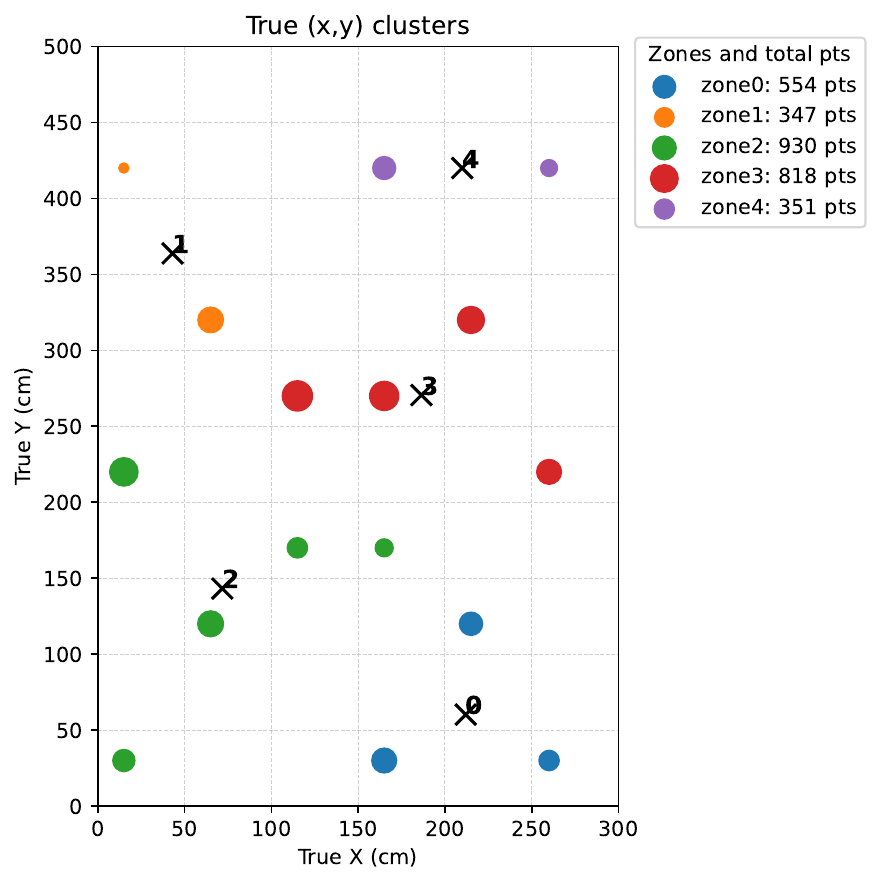}
    \includegraphics[width=0.41\linewidth]{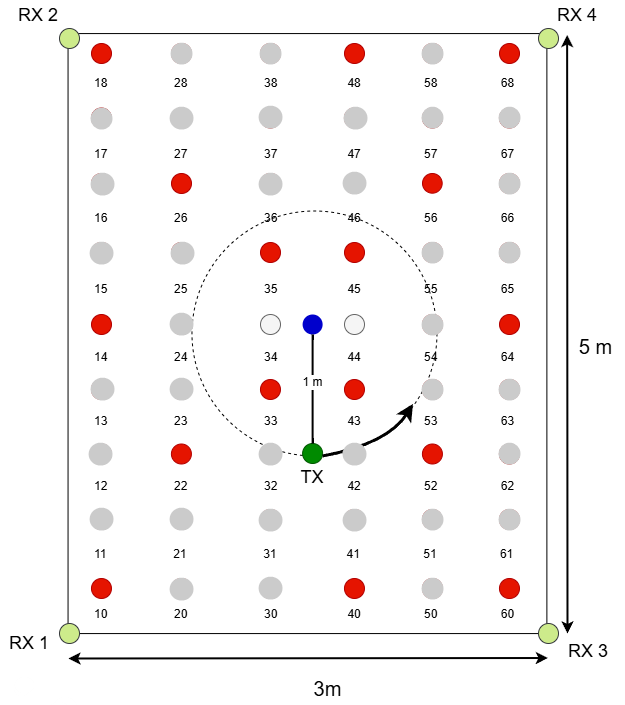}
  \caption{Left: K-means clustering of the 3000 hold-out ground-truth \((x,y)\) points into five spatial zones. Marker “\(\times\)” shows each zone centroid and area is proportional to visit count. 
    Right: Schematic of the modified experimental layout with four UWB receivers (RX 1–4), the TurtleBot-mounted transmitter (TX) on a 1 m arm, and the 16 jammer positions (red dots) in a 3 m×5 m area.}
  \label{fig:k-means}
\end{figure}

To assess whether the domain-adversarially aligned feature representations truly capture the underlying spatial structure of jammer activity, the pooled bottleneck embeddings of the $3,000$ hold-out samples were first partitioned into five zones by K-means clustering on their ground-truth $(x,y)$ coordinates. A logistic regression classifier was then trained and evaluated using only these held-out embeddings, with $5$-fold cross-validation to eliminate any bias from model adaptation. Across folds, the logistic regression achieved an average ROC-AUC of $0.9937$ and an accuracy of $0.9283$. In Figure \ref{fig:k-means}, the “$\times$” markers denote the true centroids of each zone and marker size scales with the number of visits at that location. These results confirm that a simple linear decision boundary suffices to separate spatial regions in feature space and demonstrate that the learned representations not only align across domains but also preserve fine-grained spatial discriminability.
\section{Conclusion}
\label{sec:conclusion}
The challenge of UWB jammer localization in dynamic indoor environments was addressed in this study. It was demonstrated that traditional machine learning models, while effective in a static setting (achieving a mean Euclidean error of 20.16\,cm), suffered catastrophic performance degradation when the room layout was altered, with errors increasing tenfold to 207.99\,cm. This highlighted the severe impact of domain shift on UWB localization. To overcome this, a domain-adversarial ConvNeXt autoencoder (A-CNT) was proposed. This model was shown to effectively mitigate domain shift by aligning features across different environments. A localization error of 34.67\,cm was achieved in the new environment, representing a significant 83\,\% improvement over the best source-trained baseline. Furthermore, it was shown that while other UDA methods offered benefits over source-only training, adversarial domain alignment yielded significantly greater gains. These findings underscored the necessity of adversarial learning for robust and transferable localization in dynamic, real-world settings. Future work should explore the extension of this research to 3D layouts, multiple rooms, and continuous domain adaptation.

\appendix

\section{Appendix}

\subsection{Domain Shift Visualization}

\begin{figure}[H]
    \centering
    \includegraphics[width=0.80\textwidth]{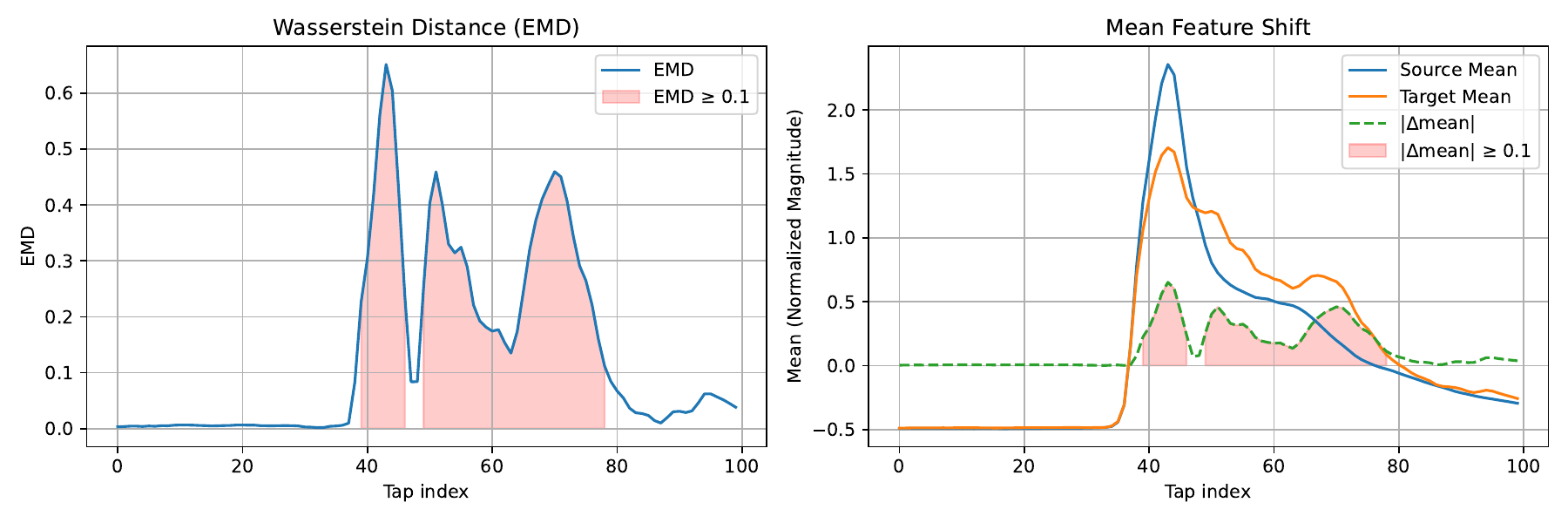}
    \caption{
        Side-by-side visualization of domain shift across taps.
        Left: Per-tap Wasserstein distance (EMD), with shaded regions indicating multiple intervals where $\text{EMD} \geq 0.1$.
        Right: Mean feature values for source and target domains, with shaded regions where the absolute difference $|\Delta\text{mean}| \geq 0.1$.
        A dashed green curve shows $|\Delta\text{mean}|$ per tap. Shaded areas are bounded by the actual metric values for visual fidelity.
    }
    \label{fig:appendix-domain-shift}
\end{figure}

\subsection{t-SNE Visualization of Feature Alignment}
\begin{figure}[H]
\centering
\includegraphics[width= 0.8\linewidth]{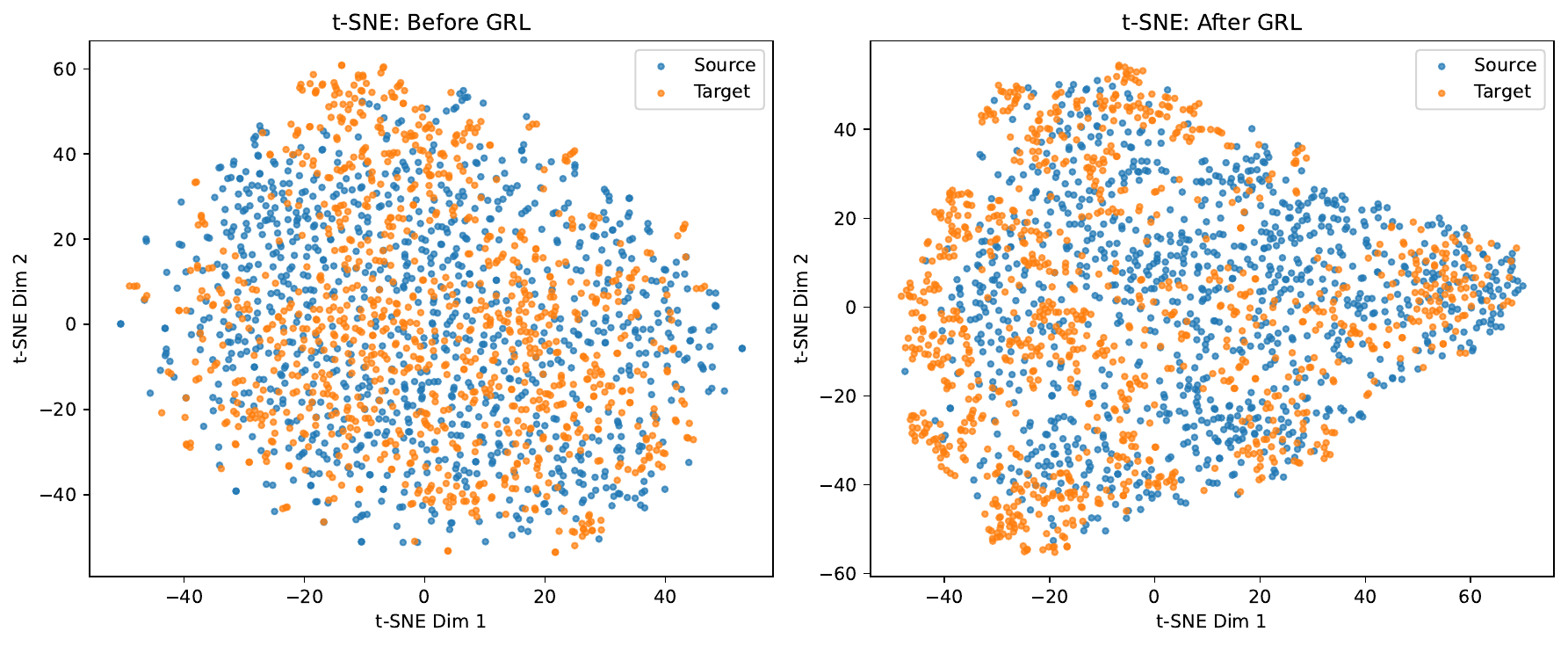}
\caption{ t-SNE visualization of feature representations before and after adversarial domain adaptation. Pretraining results in initial domain overlap, while GRL reshapes the feature space to promote domain confusion. Although overlap in the projected space is reduced after adaptation, the learned representation leads to improved target generalization, demonstrating that improved generalization can occur even as feature distributions diverge in projection.}
\label{fig:Alignment}
\end{figure}

\section*{Acknowledgments}
HF and GW were supported by the Federal Ministry of Education and Research of Germany (BMBF) within “6G-RIC: 6G Research and Innovation Cluster”, under project identification number 16KISK025, and the BMBF joint project “UltraSec: Security Architecture for UWB-based Application Platform”, project identification number 16KIS1682. GW was supported by the German Science Foundation (DFG) within priority program SPP 2378: “ResNets: Resilience in Connected Worlds” under grant WU 598/12-1.

\bibliographystyle{splncs04}
\bibliography{Paper}

\end{document}